\documentclass[11pt,a4paper]{article}
\usepackage[hyperref]{emnlp-ijcnlp-2019}
\usepackage{times}
\usepackage{latexsym}
\usepackage{url}
\usepackage{graphicx}
\usepackage{amsmath}
\usepackage{amssymb}
\usepackage{subfigure}
\usepackage{caption}
\usepackage{algorithm}
\usepackage{algorithmicx}
\usepackage{algpseudocode}
\usepackage{multicol}
\usepackage{multirow}
\usepackage{booktabs}
\usepackage{listings}
\usepackage{array}
\usepackage{comment}

\aclfinalcopy 

\title{OpenNRE: An Open and Extensible Toolkit for\\Neural Relation Extraction}

\author{Xu Han$\thanks{\quad indicates equal contribution}$\hspace{0.5em}, Tianyu Gao$^{*}$, Yuan Yao, Demin Ye, Zhiyuan Liu\thanks{\quad Corresponding author: Z.Liu(liuzy@tsinghua.edu.cn)}\hspace{0.5em}, Maosong Sun\\
Department of Computer Science and Technology, Tsinghua University, Beijing, China\\
Institute for Artificial Intelligence, Tsinghua University, Beijing, China\\
State Key Lab on Intelligent Technology and Systems, Tsinghua University, Beijing, China \\
{\tt\{hanxu17,gty16,yy18,ydm18\}@mails.tsinghua.edu.cn}
}
\date{}

\begin{document}
\maketitle
\begin{abstract}

OpenNRE is an open-source and extensible toolkit that provides a unified framework to implement neural models for relation extraction (RE). Specifically, by implementing typical RE methods, OpenNRE not only allows developers to train custom models to extract structured relational facts from the plain text but also supports quick model validation for researchers. Besides, OpenNRE provides various functional RE modules based on both TensorFlow and PyTorch to maintain sufficient modularity and extensibility, making it becomes easy to incorporate new models into the framework. Besides the toolkit, we also release an online system to meet real-time extraction without any training and deploying. Meanwhile, the online system can extract facts in various scenarios as well as aligning the extracted facts to Wikidata, which may benefit various downstream knowledge-driven applications (e.g., information retrieval and question answering). More details of the toolkit and online system can be obtained from \url{http://github.com/thunlp/OpenNRE}. 
   
\end{abstract}

\section{Introduction}

Relation extraction (RE) aims to predict relational facts from the plain text, e.g., extracting (\emph{Newton}, \texttt{the Member of}, \emph{the Royal Society}) from the sentence ``\emph{Newton} served as the president of \emph{the Royal Society}''. Because RE models can extract structured information for various downstream applications, many efforts have been devoted to researching RE. As the rapid development of deep learning in the recent years, neural relation extraction (NRE) models show the strong ability to extracting relations and achieve great performance, which makes more and more researchers and industry developers pay attention to this field.

Although the current NRE models are effective and have been applied for various scenarios, including supervised learning paradigm~\cite{zeng2014relation,nguyen2015relation,zhang2015bidirectional,zhou2016attention}, distantly supervised learning paradigm~\cite{zeng2015distant,lin2016neural,han2018hierarchical}, few-shot learning paradigm~\cite{han2018fewrel,gao2019hybrid,ye2019multi,soares2019matching,zhang2019ernie}, there still lack an effective and stable toolkit to support the implementation, deployment and evaluation of models. In fact, for other tasks related to RE, there have been already some effective and long-term maintained toolkits, such as Spacy\footnote{\url{https://spacy.io}} for named entity recognition (NER), TagMe~\cite{Tagme} for entity linking (EL), OpenKE~\cite{han2018openke} for knowledge embedding, and Stanford OpenIE \cite{angeli2015leveraging} for open information extraction. Hence, it becomes necessary and significant to systematically develop an efficient and effective toolkit for RE.

\begin{figure*}[t]
\centering
\includegraphics[width = 1.0\linewidth]{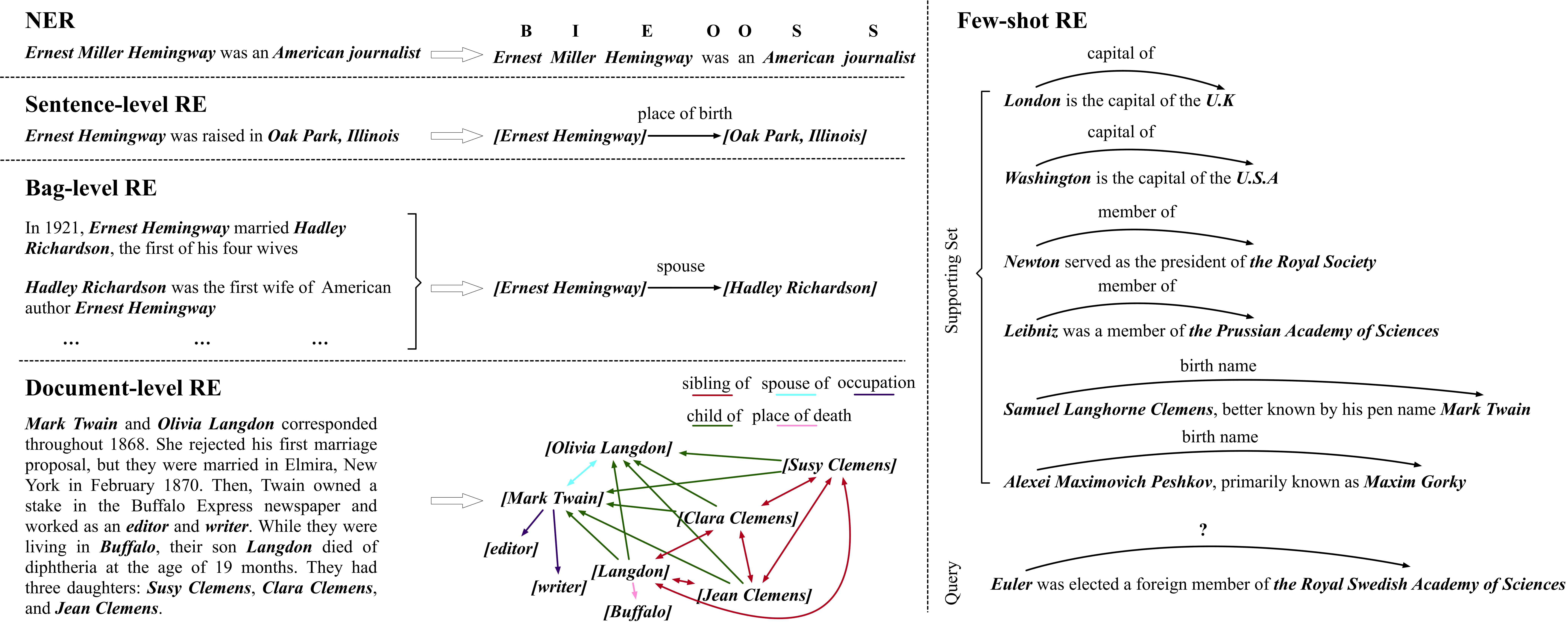}
\captionsetup{font={normalsize}}
\caption{The examples of all application scenarios in OpenNRE.}
\label{fig:example}
\vspace{-0.7em}
\end{figure*}

To this end, we develop an open and extensible toolkit for designing and implementing RE models, especially for NRE models, which is named ``OpenNRE''. The toolkit prioritizes operational efficiency based on TensorFlow and PyTorch, which support quick model training and validation. Meanwhile, the toolkit maintains sufficient system encapsulation and model extensibility, which can meet some individual requirements of incorporating new models. To keep the ease of use, we implement many typical RE models and provide a unified framework for their data processing, model training, and experimental evaluation. For those developers aiming at training custom models, they can quickly start up their RE system based on OpenNRE, without knowing too many technical details and writing tedious glue code. An online system is also available to extract structured relational facts from the text with friendly interactive interfaces and fast reaction speed. We will provide long-term maintenance to fix bugs and meet new requests for OpenNRE, and we think both researchers and industry developers can benefit from our toolkit.

\section{Application Scenarios}
\label{sec:app}

OpenNRE is designed for various scenarios for RE, including sentence-level RE, bag-level RE, document-level RE, and few-shot RE. For completing a full pipeline of extracting structured information, we also enable OpenNRE to have the capacity of entity-oriented applications to a certain extent, e.g., NER and EL. The examples of these application scenarios are all shown in Figure~\ref{fig:example}.

\subsection{Entity-Oriented Applications}

For extracting structured information from plain text, it requires to extract entities from text and then predict relations between entities. In normal RE scenarios, all entity mentions have been already annotated and RE models are just required to classify relations for all annotated entity pairs. Although the entity-oriented applications are not the focus of our toolkit, we still implement specific modules for NER~\cite{lample2016neural} and EL~\cite{han2011collective}. The NER modules can detect words or phrases (also named entity mentions) representing real-world objects. In OpenNRE, we provide two approaches for NER, one is based on spaCy, the other is based on fine-tuning BERT~\cite{devlin2018bert}. The EL modules can align those entity mentions to the entities in Wikidata~\cite{vrandevcic2014wikidata} based on TagMe~\cite{Tagme}.

\subsection{Sentence-Level Relation Extraction}

The conventional methods often handle RE in the supervised learning paradigm and extract the relation between two entities mentioned within one sentence. As shown in Figure~\ref{fig:example}, each sentence is first manually annotated with two entity mentions. Then models are required to predict the relation between those annotated entity mentions. As there are many efforts to adopt models for this setting~\cite{zeng2014relation,zhang2015bidirectional,zhou2016attention}, OpenNRE is specially designed for the sentence-level RE scenario.

\subsection{Bag-Level Relation Extraction}

The supervised RE methods suffer from several problems, especially their requirements of adequate annotated data for training. As manually labeling large amounts of data is expensive and time-consuming, \citet{mintz2009distant} introduce distant supervision to automatically label large amounts of data for RE by aligning knowledge graphs and text. Although distant supervision brings sufficient auto-labeled data, it also leads to the wrong labeling problem. Considering an entity pair may occur several times in different sentences, and there is a significant probability that some of these sentences can express the relation between the entity pair. Hence \citet{riedel2010modeling} and \citet{hoffmann2011knowledge} introduce to aggregate the sentences mentioning the same entity pair into a entity-pair bag. As shown in Figure~\ref{fig:example}, synthesizing the features of different sentences in a bag can provide more reliable information and result in more accurate predictions. The Bag-level setting is widely applied by various distantly supervised RE methods~\cite{zeng2015distant,lin2016neural,han2018hierarchical}, and thus it is also integrated into OpenNRE.

\subsection{Document-Level Relation Extraction}

\citet{yao2019docred} have pointed out that multiple entities in documents often exhibit complex inter-sentence relations rather than intra-sentence relations. Besides, as shown in Figure~\ref{fig:example}, a large number of relational facts are expressed in multiple sentences, e.g., \emph{Langdon} is the sibling of \emph{Jean Clemens}. Hence, it is hard to extract these inter-sentence relations with both the sentence-level and bag-level settings. Although the document-level RE setting is not widely explored by the current work, we argue that this scenario remains an open problem for future research, and still integrate document-level RE into OpenNRE.


\subsection{Few-Shot Relation Extraction}

Though we can train a usable and stable RE system based on the above-mentioned scenarios, which can well predict those relations appearing frequently in data, some long-tail relations with few instances in data are still neglected. Recently, some methods have been proposed to provide a different view of this problem by formalizing RE as a few-shot learning problem~\cite{han2018fewrel,gao2019hybrid,ye2019multi,soares2019matching,zhang2019ernie}. As shown in Figure~\ref{fig:example}, each relation only have a handful of instances in the supporting set in a few-shot RE scenario, and models are required to be capable of accurately capturing relation patterns of these small amounts of training instances. Considering few-shot RE is important for handling long-tail relations, OpenNRE also provides a custom platform for further research in this direction.

\begin{figure}[t]
    \centering
    \includegraphics[width=0.49\textwidth]{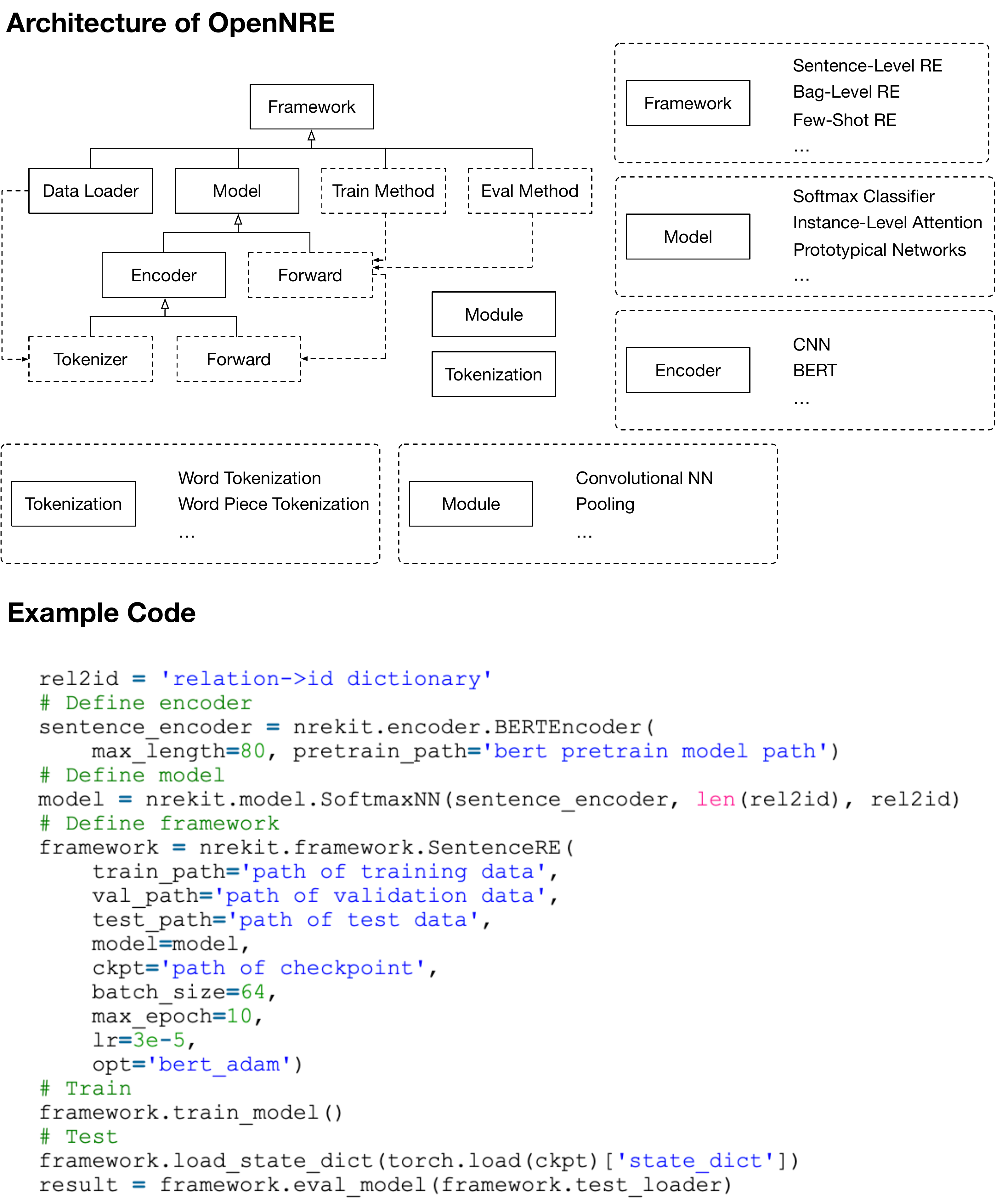}
    \caption{The architecture and example code of OpenNRE. The structure shows the contents of each part of OpenNRE and how they are related. Based on OpenNRE, one can use only a few lines of code to define, train and evaluate RE models of different scenarios.}
    \label{fig:my_label}
    \vspace{-0.7em}
\end{figure}

\section{Toolkit Design and Implementation}

Our goal of designing OpenNRE is achieving the balance among system encapsulation, operational efficiency, model extensibility, and ease of use. 

For system encapsulation, we build a unified underlying platform to encapsulate various data processing and training strategies, so that developers can maximize the reuse of code to avoid unnecessary redundant model implementations. For operational efficiency, OpenNRE is based on TensorFlow and PyTorch, which enables developers to train models on GPUs. For model extensibility, we systematically implement various neural modules and some special algorithms (e.g., adversarial training~\cite{wu2017adversarial} and reinforcement learning~\cite{feng2018reinforcement}). Hence, it is easy to implement new RE models based on OpenNRE. We also implement some typical RE models so as to conveniently train custom models for specific application scenarios.

More specifically, OpenNRE attains the above four design objects through implementing the following five components.

\subsection{Tokenization}

The tokenization component is responsible for tokenizing input text into several input tokens. In OpenNRE, we implement both word-level tokenization and subword-level tokenization. These two operations satisfy most tokenization demands and help developers get rid of spending too much time writing glue code for data processing. For building a new tokenizer, extending the \texttt{BasicTokenizer} class and implementing specific tokenization operations is convenient.

\subsection{Module}

The module component consists of various functional neural modules for model implementation, such as the essential network layers, some pooling operations, and activation functions. In order to adapt these modules for RE scenarios, we also implement some special RE neural modules (e.g., piece-wise pooling operation~\cite{zeng2015distant}). Using these atomic modules to construct and deploy RE systems has a high degree of freedom.

\subsection{Encoder}

The encoder is applied to encode text into its corresponding embeddings to provide semantic features. In OpenNRE, we implement the \texttt{BaseEncoder} class based on the tokenization and module components, which can provide basic functions of text tokenization and embedding lookup. By extending the \texttt{BaseEncoder} class and implementing specific neural encoding architecture, we can implement various specific encoders. In OpenNRE, we have implemented the common convolutional and recurrent neural encoders, as well as the pre-trained encoder BERT.

\subsection{Model}

Some developers may not require to implement and verify their own RE models, and their main demand is to easily train and deploy custom models. To this end, we also replicate several typical RE models~\cite{zeng2015distant,zhang2015bidirectional}. Some special algorithms for enhancing RE models are also included in the toolkit, such as attention mechanism~\cite{lin2016neural}, adversarial training~\cite{wu2017adversarial}, and reinforcement learning~\cite{feng2018reinforcement}. On the one hand, the model component enables us to train custom models without having to understand all technical details. On the other hand, the implemented models in this model component are all tutorial examples to show how to build models with OpenNRE.

\subsection{Framework}

The framework module is mainly responsible for integrating other four components and supporting various functions (including data processing, model training, model optimizing, and model evaluating). In OpenNRE, for all application scenarios mentioned in Section~\ref{sec:app}, we have implemented their corresponding framework. For other future potential application scenarios, we have also reserved interfaces for their implementation.

\begin{table}[t]
    \centering
    \small
    \begin{tabular}{c|c|c}
      \toprule
        Model & Wiki80 & SemEval\\
        \midrule
        CNN  & $63.93$ & $71.11$  \\
        BERT & $84.57$ & $84.02$  \\
        BERT-Entity & $86.61$ & $84.21$\\
      \bottomrule
    \end{tabular}
    \caption{Accuracies of various models on Wiki80 and SemEval 2010 Task-8 under the single sentence setting.}
    \label{tab:wiki80}
    \vspace{-0.7em}
\end{table}

\begin{table}[t]
    \centering
    \small
    \begin{tabular}{c|c|c}
      \toprule
       Model & F1  &  F1 (*)\\
       \midrule
        BERT & $0.880$ & - \\
        BERT-Entity &  $0.883$ & $0.892$ \\
      \bottomrule
    \end{tabular}
    \caption{Micro F1 scores of various models on SemEval 2010 Task-8 under the sentence-level RE setting. ``(*)'' indicates the original results from \citet{soares2019matching}.}
    \label{tab:semevalf1}
    \vspace{-0.7em}
\end{table}

\section{Experiment and Evaluation}

In this section, we evaluate our toolkit on several benchmark datasets in different RE scenarios. The evaluation results show that our implementation of some state-of-the-art models with OpenNRE can achieve comparable or even better performance, as compared to the original papers.

\subsection{Sentence-Level Relation Extraction}

\label{sec:exp_sent}

\begin{table*}[t]
    \centering
    \small
    \begin{tabular}{c|c|c|c|c}
      \toprule
        Model & 5-Way 1-Shot & 5-Way 5-Shot& 5-Way 1-Shot (*) & 5-Way 5-Shot (*)\\
      \midrule
        Prototype-CNN & $74.5$ & $88.4$ & $69.2$& $84.8$\\
        Prototype-BERT & $80.7$ & $89.6$ & - & -\\
        BERT-PAIR & $88.3$ & $93.2$ & - & - \\
      \bottomrule
    \end{tabular}
    \caption{Accuracies of various models on FewRel under the different few-shot settings. ``(*)'' indicates the original results taken from~\citet{han2018fewrel}.}
    \label{tab:fewrel}
    \vspace{-0.7em}
\end{table*}

\begin{table}[t]
    \centering
    \small
    \begin{tabular}{c|c|c|c|c}
        \toprule
        Model & AUC & F1 & AUC (*) & F1 (*)\\
        \midrule
        CNN-ATT & $0.333$  & $0.397$  & $0.318$ & $0.380$\\
        CNN-ADV & $0.337$  & $0.406$  & - & -\\
        CNN-RL  & $0.276$  & $0.429$  & - & $0.42$\\
        \bottomrule
    \end{tabular}
    \caption{AUC and F1 scores of various models on NYT10 under the bag-level RE setting. ``(*)'' indicates the original results taken from~\citet{lin2016neural} and~\citet{feng2018reinforcement}.}
    \label{tab:nyt10}
    \vspace{-0.6em}
\end{table}

We experiment on two different encoders for sentence-level relation extraction: CNN \cite{zengrelation} and BERT \cite{devlin2018bert}. For CNN, we follow the setting of \citet{nguyen2015relation}, including using word and position embeddings. For BERT, we follow the setting of \citet{soares2019matching}. ``BERT" in our paper refers to using entity markers in input and taking \texttt{[CLS]} as output. ``BERT-Entity" refers to using entity markers in input and taking entity start as output like \citet{soares2019matching}.

We carry out experiments on two datasets of sentence-level RE: SemEval 2010 Task-8 \cite{hendrickx2009semeval} and Wiki80. SemEval 2010 Task-8 contains 19 relations and 10,717 instances, $17.4\%$ of which are with no relation. Wiki80 is derived from FewRel \cite{han2018fewrel}, a large scale few-shot dataset. It contains 80 relations and 56,000 instances from Wikipedia and Wikidata \cite{vrandevcic2014wikidata}. Since Wiki80 is not an official benchmark, we directly report the results on the validation set. From Table \ref{tab:wiki80} we can see that BERT-based models perform better than the CNN model and achieve promising results on both datasets. Our implementation of BERT-Entity with OpenNRE achieves comparable results to the original work in Table \ref{tab:semevalf1}.

\subsection{Bag-Level Relation Extraction}

We implement models with instance-level attention \cite{lin2016neural}, adversarial training \cite{wu2017adversarial} and reinforcement learning \cite{feng2018reinforcement} for bag-level relation extraction. Note that the latter two are based on the instance-level attention mechanism. We use the same CNN encoder as Section \ref{sec:exp_sent} and denote the three models as ``CNN-ATT'', ``CNN-ADV'' and ``CNN-RL''. We evaluate those three models on NYT10 \cite{riedel2010modeling}, a distantly supervised dataset based on New York Times corpus and FreeBase \cite{bollacker2008freebase}. Table \ref{tab:nyt10} shows that our version of bag-level RE models achieves comparable or even better results than the original papers. 

\subsection{Few-Shot Relation Extraction}

We experiment on Prototypical Networks \cite{snell2017prototypical} for few-shot RE. For the encoder selection, we take CNN and BERT as described in Section \ref{sec:exp_sent}. We also implement a model named ``BERT-PAIR'', which takes one supporting sentence and one query sentence as input, and directly outputs the probability that they share the same relation with the BERT sequence classification model. The experiments are carried out on FewRel described in Section \ref{sec:exp_sent}. From Table \ref{tab:fewrel} we can see that for both few-shot settings, our version achieves better results than the original results from \citet{han2018fewrel}, proving that our implementation with OpenNRE is robust.

\section{Online System}

\begin{figure}[t]
\centering
\includegraphics[width = 1.0\linewidth]{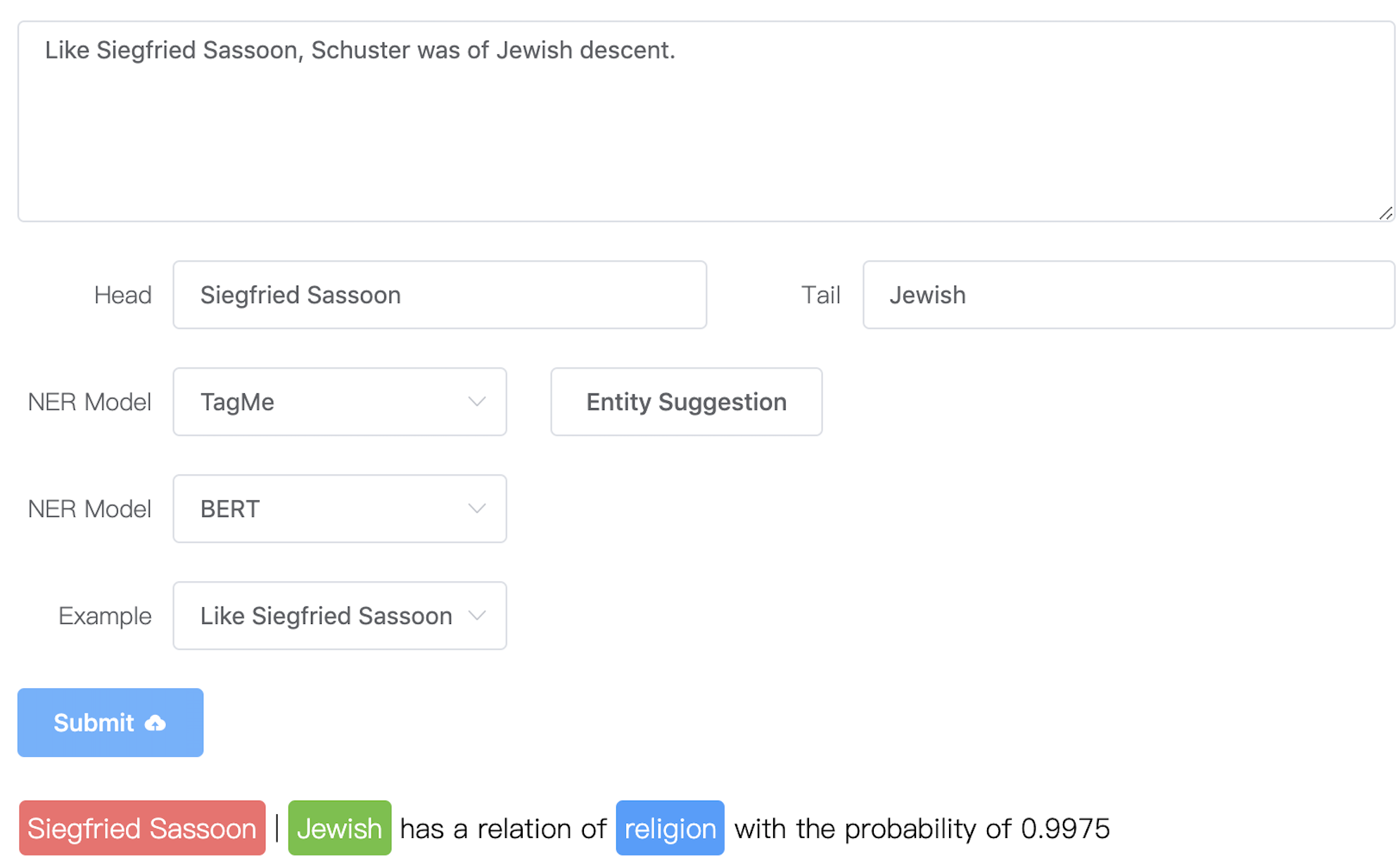}
\captionsetup{font={normalsize}}
\caption{An example of the online system.}
\label{fig:demo}
\vspace{-0.6em}
\end{figure}

Besides the toolkit, we also release an online system. As shown in Figure~\ref{fig:demo}, we train a model in the sentence-level RE scenario and deploy the model for online access. The online system can be directly applied for extracting structured facts from plain text. Meanwhile, all extracted entity mentions and relations can be aligned to Wikidata.

\section{Conclusion}

We propose OpenNRE, an open and extensible toolkit for relation extraction. OpenNRE achieves the balance among system encapsulation, operational efficiency, model extensibility, and ease of use. Based on OpenNRE, either training custom models or quick model validation becomes easy. Some experimental results also demonstrate that the models implemented by OpenNRE are efficient and effective, which can achieve comparable or even better performance as compared to the original papers. Furthermore, an online system is also available for meeting real-time extraction without training and deploying. In the future, we will provide long-term maintenance to fix bugs and meet new requests.

\section*{Acknowledgments}

This work is supported by the National Key Research and Development Program of China (No. 2018YFB1004503) and the National Natural Science Foundation of China (NSFC No. 61572273, 61772302). Han and Gao are supported by 2018 and 2019 Tencent Rhino-Bird Elite Training Program respectively. Gao is also supported by Tsinghua University Initiative Scientific Research Program.

\small
\bibliography{emnlp-ijcnlp-2019}
\bibliographystyle{acl_natbib}

\end{document}